\documentclass{bmvc2k}

\title{Very Efficient Training of Convolutional Neural Networks using Fast Fourier Transform and Overlap-and-Add}

\addauthor{Tyler Highlander$^*$}{highlander.2@wright.edu}{1}
\addauthor{Andres Rodriguez$^{*,\star}$}{andres.rodriguez.8@us.af.mil}{2}

\addinstitution{
 $^*$College of Eng. and Comp. Science\\
 Wright State University\\
 Dayton, Ohio, USA
}
\addinstitution{
 $^\star$Air Force Research Laboratory\\
 Dayton, Ohio, USA
}

\runninghead{Highlander, Rodriguez}{Efficient Training of CNNs using FFT and OaA}

\def\etal{\emph{et al}\bmvaOneDot}
\begin{document}

\maketitle

\begin{abstract}
Convolutional neural networks (CNNs) are currently state-of-the-art for various classification tasks, but are computationally expensive. Propagating through the convolutional layers is very slow, as each kernel in each layer must sequentially calculate many dot products for a single forward and backward propagation which equates to $\mathcal{O}(N^{2}n^{2})$ per kernel per layer where the inputs are $N \times N$ arrays and the kernels are $n \times n$ arrays. Convolution can be efficiently performed as a Hadamard product in the frequency domain. The bottleneck is the transformation which has a cost of $\mathcal{O}(N^{2}\log_2 N)$ using the fast Fourier transform (FFT). However, the increase in efficiency is less significant when $N\gg n$ as is the case in CNNs. We mitigate this by using the ``overlap-and-add'' technique reducing the computational complexity to $\mathcal{O}(N^2\log_2 n)$ per kernel. This method increases the algorithm's efficiency in both the forward and backward propagation, reducing the training and testing time for CNNs. Our empirical results show our method reduces computational time by a factor of up to 16.3 times the traditional convolution implementation for a 8 $\times$ 8 kernel and a 224 $\times$ 224 image.
\end{abstract}

\section{Introduction}
\label{sec:intro}
Convolutional neural networks (CNNs) achieved state-of-the-art classification rates on various datasets \cite{deng2009imagenet, lecun1998mnist, krizhevsky2009learning}, but require significant computational resources. For example, AlexNet \cite{krizhevsky2012imagenet} has over 60 million free parameters trained with stochastic gradient descent requiring thousands of forward and backward propagations through a network with 5 convolutional layers. A more recent CNN, GoogLeNet \cite{szegedy2015going}, has various layers within layers amounting to 59 convolutional layers.  Propagating through these convolutional layers is the computational bottleneck of training and testing CNNs. Standard convolutional layers are slow, as each kernel must calculate many dot products for a single forward and backward propagation which equates to $\mathcal{O}(N^{2}n^{2})$ per kernel, where the inputs are $N \times N$ arrays and the kernels are $n \times n$ arrays. To converge to a local minimum, CNNs usually require hundreds of epochs. An epoch consists of propagating all the training samples in the dataset through the network once. In addition, it is common to train multiple CNNs for one task and compute an average of the multiple outputs in testing. With over one million training images in the ImageNet Large Scale Visual Recognition Challenge (ILSVRC) \cite{deng2009imagenet} and hundreds of epochs needed for training each CNN, reducing the complexity of the convolution operation reduces training time.

Convolution can be efficiently computed in the frequency domain as a Hadamard product. The computational bottleneck is the Fourier transform between the space and the frequency domain. For an input of size $N^2$, this  can be efficiently computed using fast Fourier transforms (FFTs) with complexity $\mathcal{O}(N^2 \log_2 N)$. Mathieu \etal \cite{mathieu2013fast} demonstrated that this reduces the training and testing time of CNNs. In their work they efficiently calculated FFTs on a GPU and used these transforms to perform convolutions via a Hadamard product in the frequency domain.

In this paper, we propose to use the overlap-and-add (OaA) technique \cite{oppenheim1989discrete} to further reduce the training and testing complexity to $\mathcal{O}(N^2\log_2 n)$ per kernel. Note that the overlap-and-save \cite{oppenheim1989discrete} is a similar technique that may be marginally faster but has the same complexity. In a CNN convolutional layer, the input array and the set of $K$ kernel arrays have a depth of size $C$, e.g., $C=3$ for an RBG input image. The total number of convolutions in a CNN convolutional layer is $KC$; each channel of each kernel is convolved with the respective channel in the input array. For each of these convolutions, OaA should be used to improve efficiency with no cost in performance. Section 2 explains the OaA technique and our convolution implementation. In Section 3 we demonstrate that our method computationally outperforms (by a factor of up to 16.3) traditional implementations. We offer concluding remarks in Section 4.

\section{Overlap-and-Add}
\begin{figure}[!htbp]
\caption{1-D Overlap-and-Add convolutions}
  \centering
    \includegraphics[width=1.0\textwidth]{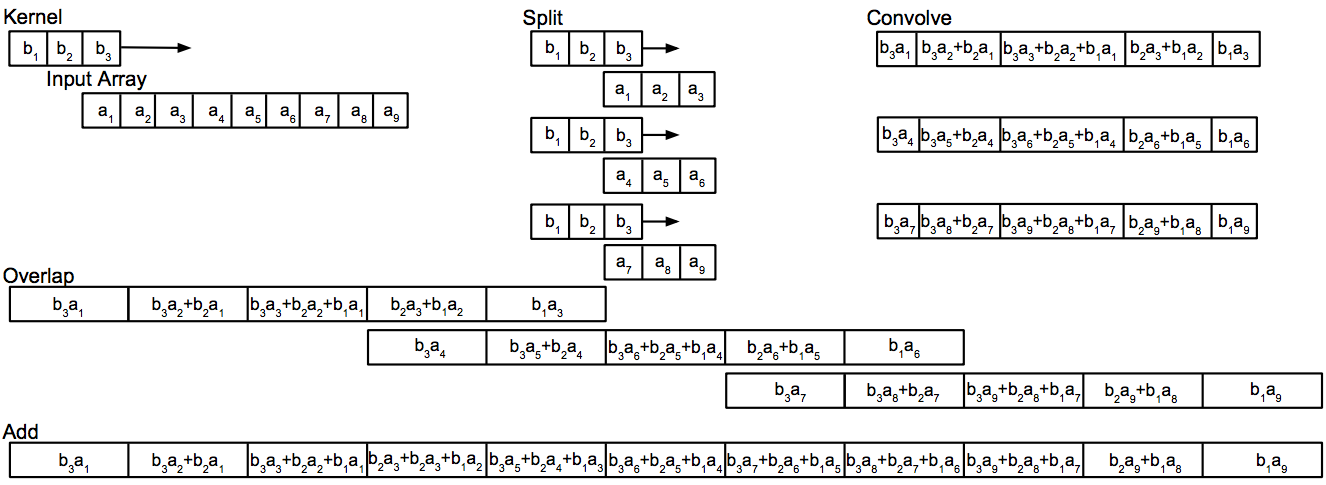}
    \label{fig:OaA}
\end{figure}

In OaA, the input is broken into ${N^2}/{n^2}$ (rounded up) blocks equal to the kernel size $n\times n$. A convolution between each block and the kernel is computed and the results are overlapped and added. Figure \ref{fig:OaA} illustrates a simple 1-D overlap-and-add method for spacial convolution (that can easily generalize to 2-D). The input array is first split into smaller blocks that are the size of the kernel. Smaller convolutions are computed between the kernel and the block inputs. The resulting convolutions are overlapped by $n-1$, where $n$ is the length of the kernel, and added together to create the same results as a traditional spacial convolution.

Each convolution in OaA can be efficiently computed in the frequency domain, where the bottleneck is the complexity of each 2-D fast Fourier transform $\mathcal{O}(n^2 \log_2 n)$. The total complexity for the entire input and kernel is the number of blocks times the complexity of each block convolution, i.e., $\mathcal{O}(N^2 \log_2 n)$. Table \ref{tab:complexity} compares the complexity of each method, where spaceConv refers to the traditional convolution in the space domain, FFTconv refers to convolution via a Hadamard product in the frequency domain without overlap-and-add, and OaAconv refers to convolution using overlap-and-save where each smaller convolution is efficiently computed in the frequency domain.

\begin{table}[!htbp]
\centering
  \begin{tabular}{ | c | c | }
    \hline
     Method & Computational Complexity \\ \hline
    spaceConv & $\mathcal{O}(N^{2}n^{2})$ \\ \hline
    FFTconv & $\mathcal{O}(N^{2}log_2 N)$ \\ \hline
     \textbf{OaAconv} & {$\mathcal{O}(N^{2}log_2 n)$} \\
    \hline
  \end{tabular}
  \caption{Computational Complexity Comparison}
  \label{tab:complexity}
\end{table}

When $N\gg n$ as is the common case in CNN architectures, OaAconv reduces the computational complexity by a factor of $\frac{n^2}{\log_2 n}$ over spaceConv and by a factor of $\frac{\log N}{\log n}$ over FFTconv. For example, for a $256\times 256$ input array with a $5\times 5$ kernel (typical values in a CNN architecture), spaceConv has a complexity of $\mathcal{O}(256^2 \times 25)$, FFTconv has a complexity of $\mathcal{O}(256^2 \times 8)$, and OaAconv has a complexity of $\mathcal{O}(256^2 \times 2.3)$.

The overall time complexity of OaAconv can be further reduced by noting that all the block convolutions can be computed in parallel. If $N^2/n^2$ threads are available (a fair assumption for modern GPUs), the complexity on each thread is $n^2\log_2 n$, and the overall time complexity is $\mathcal{O}(\max(N^2, n^2\log_2 n))$ which is usually $\mathcal{O}(N^2)$. We can get additional speed up by taking advantage of the NVIDIA CUDA Fast Fourier Transform library (cuFFT) that computes each individual FFTs up to 10 times faster \cite{nvidia2010cufft} (see also \cite{vasilache2014fast}). However, in order to have a fair comparison, our experiments in this paper are run on single threads. As part of this work, we created a Caffe \cite{jia2014caffe} fork\footnote{github.com/THighlander} that uses a multi-thread GPU implementation of OaA for efficient convolutions.

A possible area of concern for OaA is the additional cost of breaking the input into blocks and the overlapping and adding cost after each convolution. Our experiments show that the performance increase of the OaA technique outweigh these overhead costs.

It is worth noting that although OaA always reduces the computational complexity in testing (and training), it is particularly beneficial in implementations such as Sermanet \etal \cite{sermanet2013overfeat} when the test image is much larger than the training images, and uses a sliding window approach across a pyramid of scales for simultaneous detection (localization) and classification.

\section{Experiments and Results}

\subsection{Training consistency}

In this experiment we use each method of convolutions to train the CNN LeNet-5 \cite{lecun1995convolutional} architecture using the MNIST \cite{lecun1998mnist} dataset. The goal of this experiment is to show empirically that the methods are equivalent. Each network is trained for only 100 epochs as all we need to show is consistency. Each CNN network is trained five times with each type of convolution technique, and their classification rate averages are shown in Table \ref{tab:avgrates}.
\begin{table}[!htbp]
\centering
  \begin{tabular}{ | c | c | }
    \hline
     Convolution method & Performance rate (averaged over 5 networks) \\ \hline
    spaceConv & 92.48\% \\ \hline
    FFTconv & 92.41\% \\ \hline
    \textbf{OaAconv} & 92.46\% \\
    \hline
  \end{tabular}
  \caption{Performance rates for networks trained with various convolution methods}
  \label{tab:avgrates}
\end{table}

As expected, all three methods averaged within 0.07\% of each other. The reason there is a non-zero difference is the random parameter initialization in training each CNN.

\subsection{Time vs. number of kernels}
\begin{figure}[ht!]
  \caption{Speed-up over spaceConv vs. number of kernels in forward propagation}
  \centering
  \includegraphics[width=1.0\textwidth]{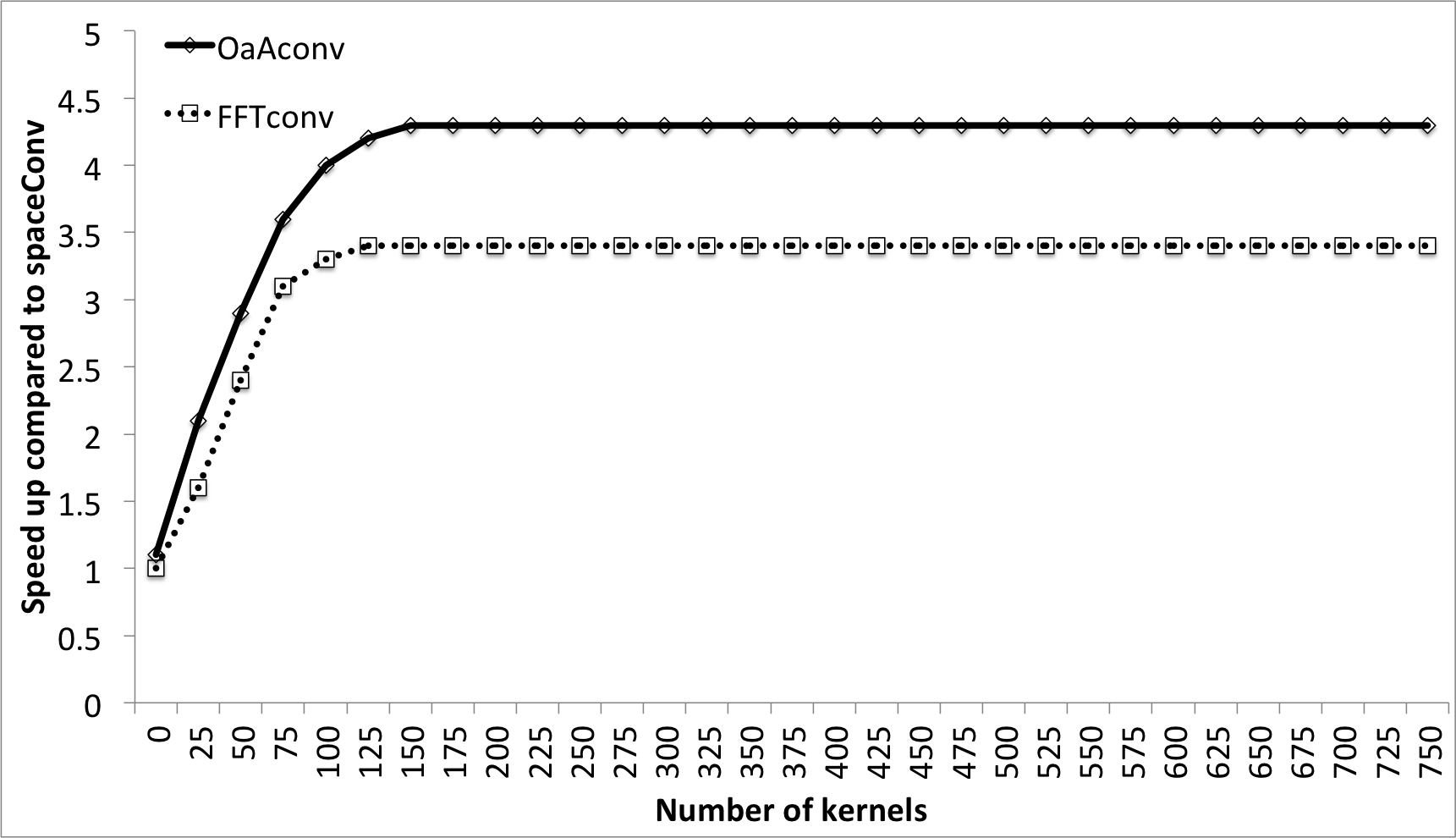}
  \label{fig:numKvsFP}
  \setlength{\belowcaptionskip}{15pt}
\end{figure}

In this experiment we compare the required total propagation time through one convolutional layer as the number of kernels in the layer increases. To compare computational time, note that additional channels in the CNN convolutional layer input array can be treated as a multiplicative factor in the number of kernels, i.e., the computations required to convolve a $C$-channels input array with a set of $K$-channels kernels is equivalent to convolving a 1-channel input array with a set of 1-channel $KC$ kernels. In our experiments, the input array is of size 32 $\times$ 32 and each kernel is of size 5 $\times$ 5, both with 1-channel. Figure \ref{fig:numKvsFP} shows the speed-up factor of FFTconv and OaAconv compared to spaceConv in the forward propagation of the convolutional layer as the number of kernels varies. The number of kernels is varied from 25 to 750 with a discrete step of 25. Each ``number of kernels'' experiment is repeated 10 times and the results are averaged.

\vspace{+5mm}

FFTconv and OaAconv outperform spaceConv, and OaAconv outperforms FFTconv at every step. FFTconv and OaAconv have an additional initialization cost: in OaA the input array must be divided, and in both methods the input (or divided input) array and kernel must be zero-padded, so that each side is $N+n-1$ in FFTconv and $2n-1$ in OaAconv, prior to computing the Fourier transforms. As the number of kernels increases, these additional initialization costs becomes less significant.

To quantify the initialization cost of FFTconv and OaAconv, we convolved an input array of size 224 $\times$ 224 with a kernel of size 8 $\times$ 8. This experiment is repeated 10 times and averaged. OaAconv spends 1.6\% of its total calculations on handling the overhead, while FFTconv spends 8.2\%. In our implementation spaceConv does not require any overhead. The large difference between the input size and kernel size causes FFTconv to have a larger overhead than OaAconv.

\begin{figure}[ht!]
  \caption{Speed-up over spaceConv vs. number of kernels in backward propagation}
  \centering
    \includegraphics[width=1.0\textwidth]{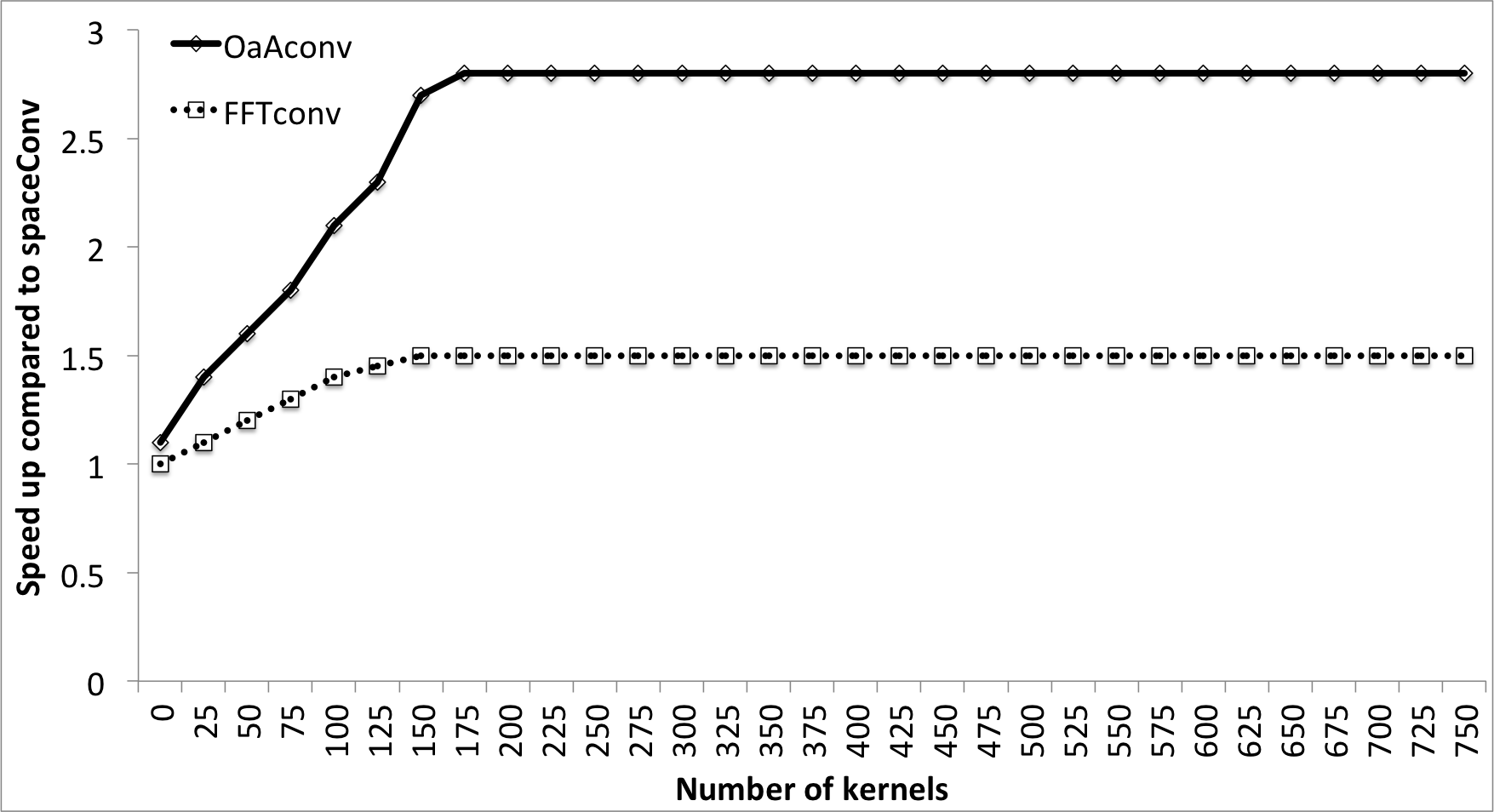}
    \label{fig:numKvsBP}
\end{figure}

Figure \ref{fig:numKvsBP} shows the speed-up over spaceConv vs. number of kernels for the backward propagation. Our method outperforms FFTconv more than in the forward propagation. The reason for this is that the backward propagation contains two actual convolutions per kernel: one convolution to propagate the error through the layer and another to calculate the change in weight generated by this error.

\vspace{-5mm}
These experiments show that the additional initialization costs of using OaAconv and FFTconv are mitigated by the lower complexity of these methods. The more kernels used, the larger the performance increase of our method.

\subsection{Time vs. kernel size}

\begin{figure}[ht!]
  \caption{Speed-up over spaceConv vs. kernel size in forward propagation}
  \centering
    \includegraphics[width=1.0\textwidth]{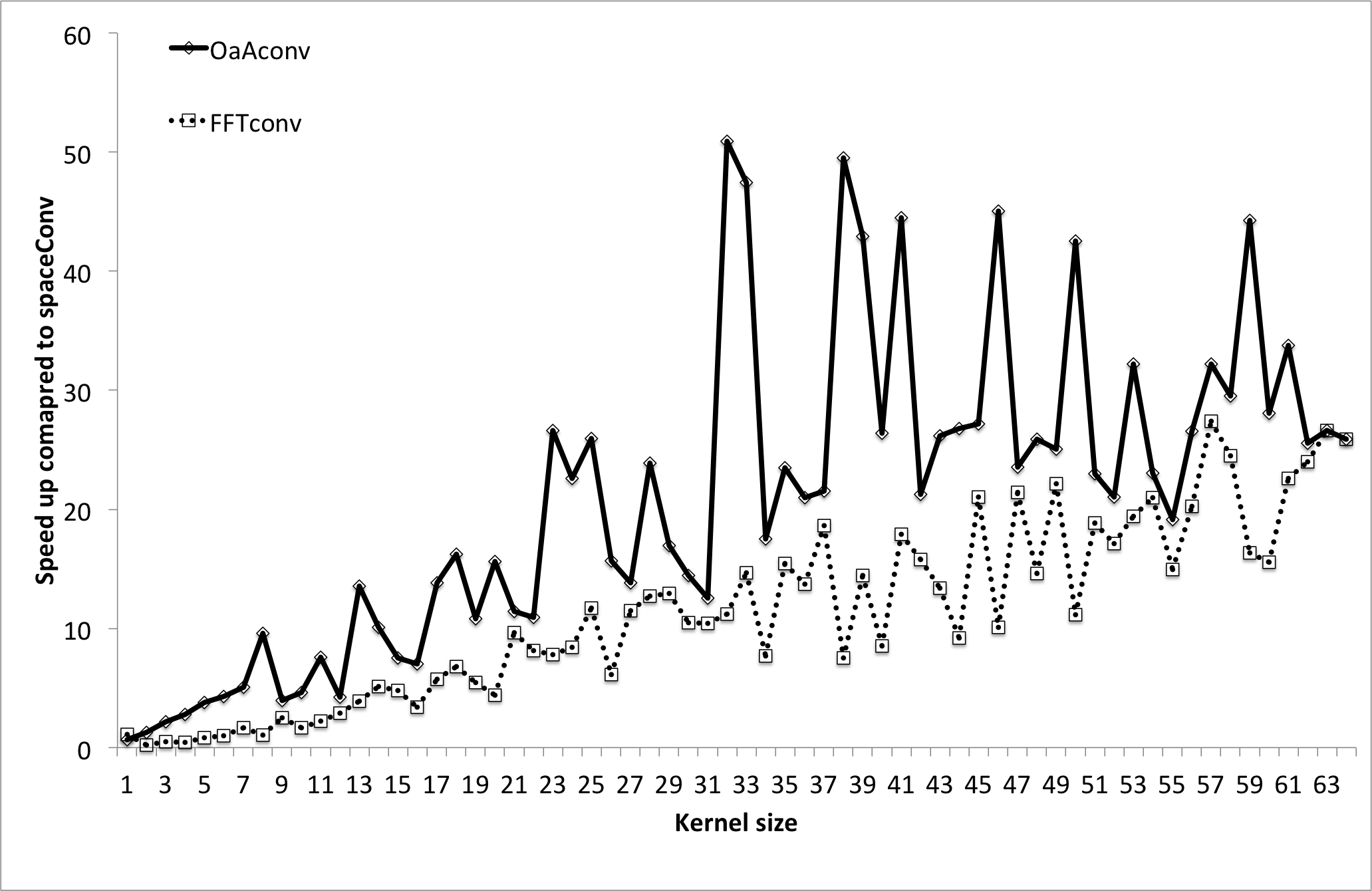}
    \label{fig:sizeKvsFP}
\end{figure}

\begin{figure}[ht!]
  \caption{Speed-up over spaceConv vs. kernel size in backward propagation}
  \centering
    \includegraphics[width=1.0\textwidth]{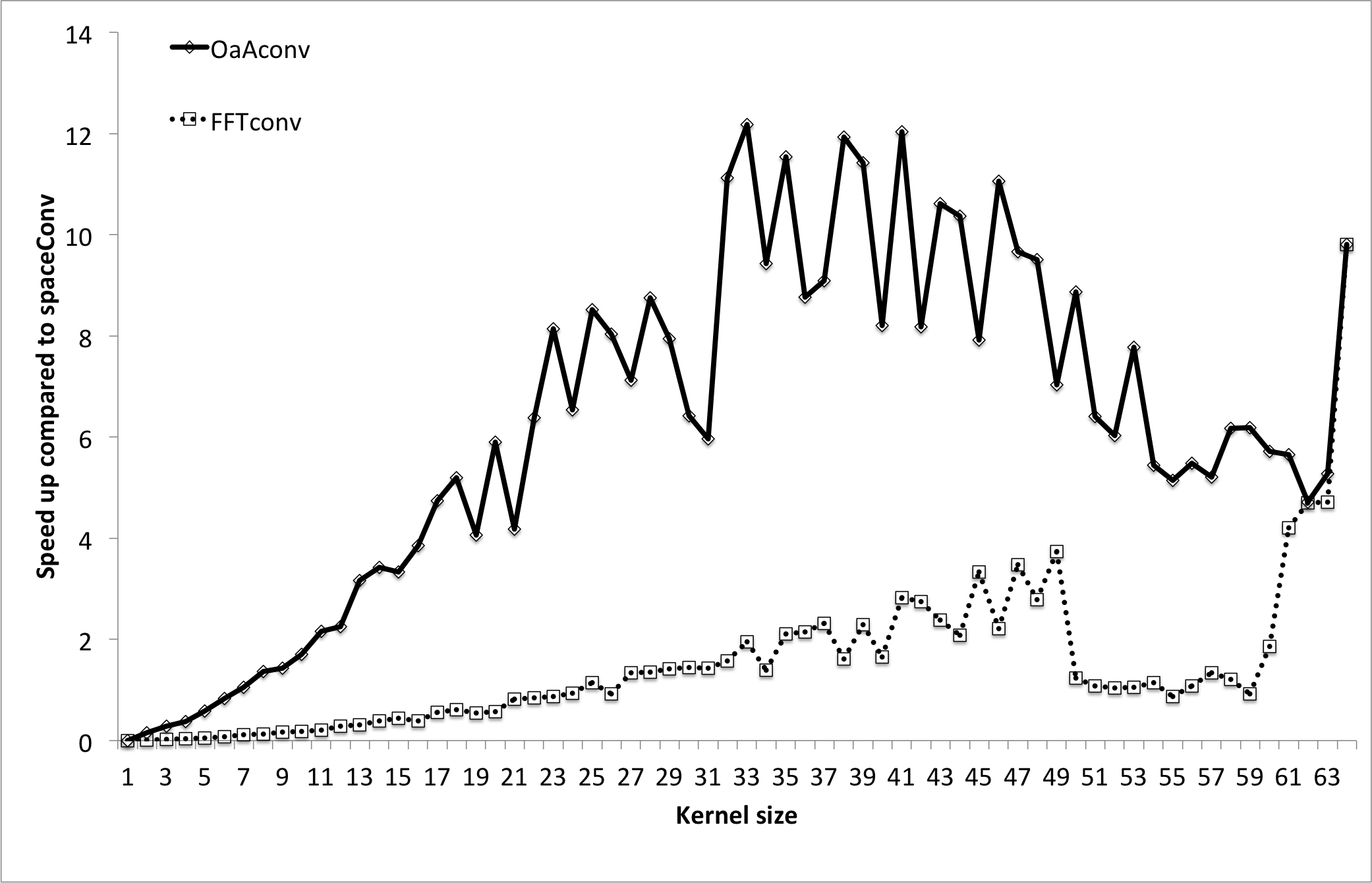}
    \label{fig:sizeKvsBP}
\end{figure}

In this experiment we vary the size of the kernel while keeping the input size constant. The size of the input array is 64 $\times$ 64. The number of kernels used is held constant at 100. Figures \ref{fig:sizeKvsFP} and \ref{fig:sizeKvsBP} show the speed-up over spaceConv vs. kernel size for the forward and backward propagation, respectively. The kernel sizes vary from 1 to 64 with a discrete step of 1. Each ``kernel size'' experiment is repeated 10 times and the results are averaged.

\vspace{+5mm}
\vspace{+5mm}
\vspace{+5mm}
In the forward propagation in Figure \ref{fig:sizeKvsFP}, the performance of FFTconv and OaAconv converge at 64 as expected. An interesting aspect of this graph is the various performance peaks at different kernel sizes. This is because the FFT software \cite{frigo1998fftw} is optimal for Fourier transforms with power of 2 sides along each dimension. We can leverage this fact when designing CNN architectures to further reduce computational requirements, and/or we can zero-pad to the nearest power of 2 to complement the design of the FFT algorithm. Note that the backward propagation has peak performances at different kernel sizes due to different zero-padding sizes prior to the Fourier transform.

\subsection{Time vs. input size}

\begin{figure}[ht!]
\caption{Speed-up over spaceConv vs. input size in forward propagation}
  \centering
    \includegraphics[width=1.0\textwidth]{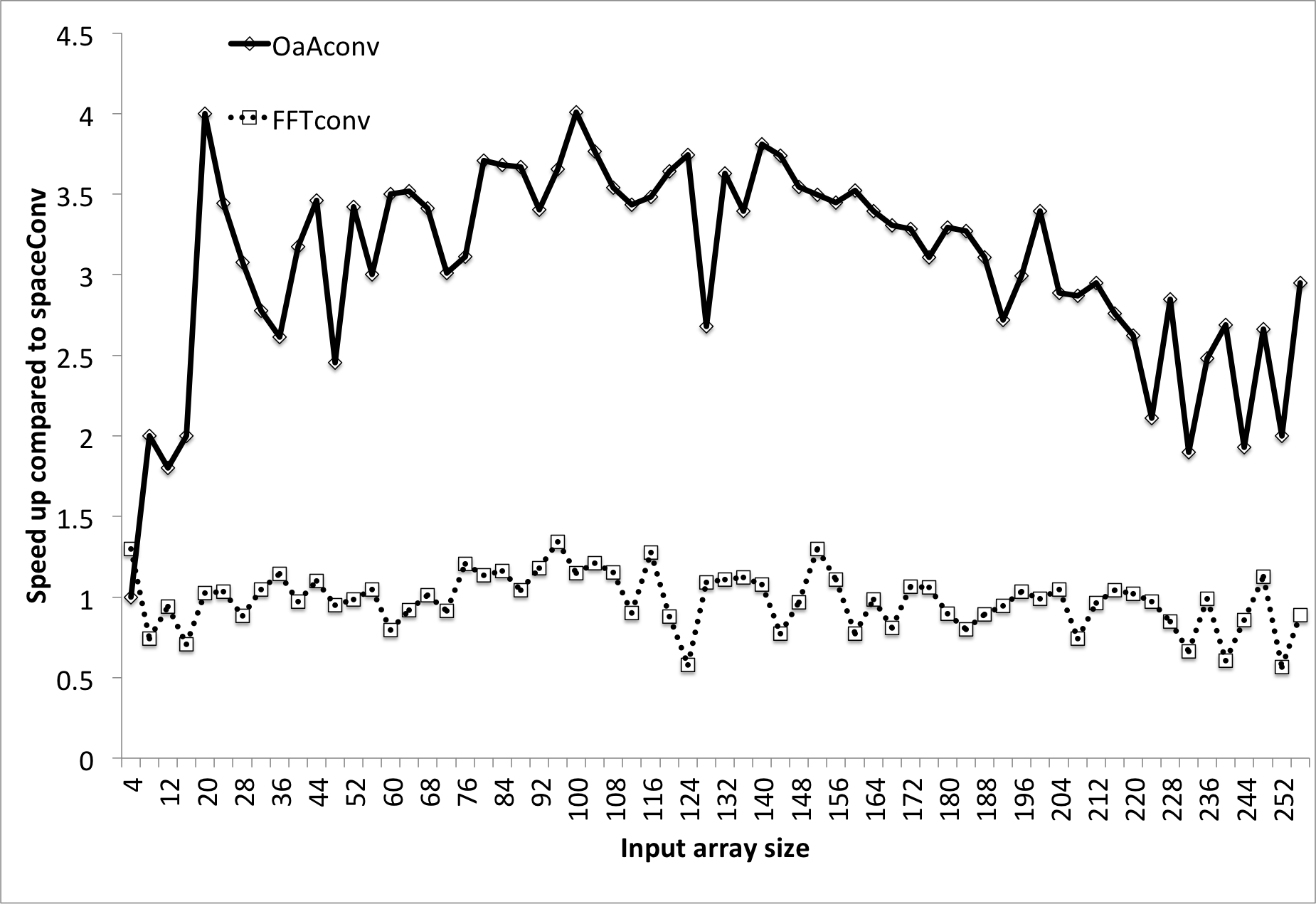}
    \label{fig:imsizevsFP}
\end{figure}

\begin{figure}[ht!]
\caption{Speed-up over spaceConv vs. input size in backward propagation}
  \centering
    \includegraphics[width=1.0\textwidth]{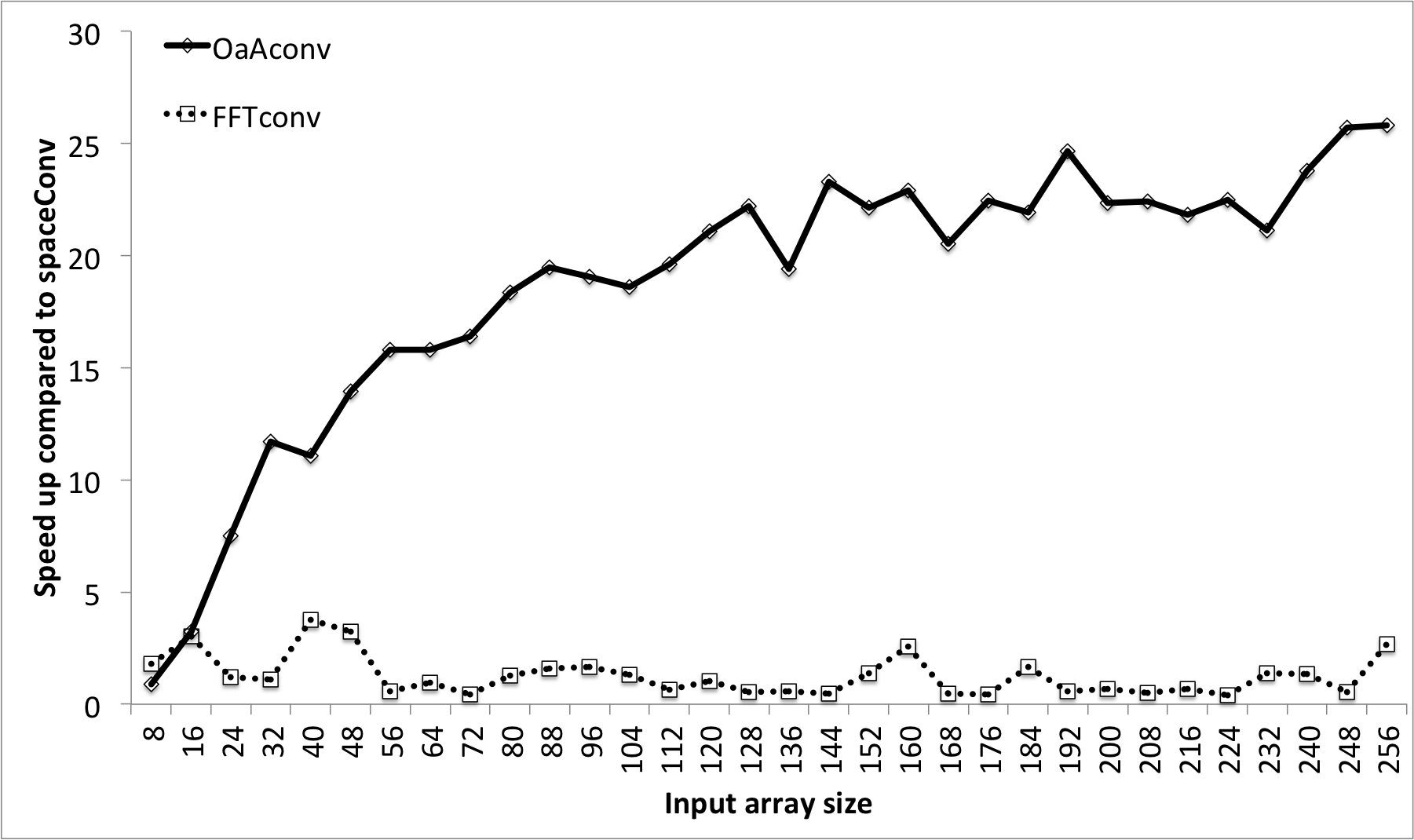}
    \label{fig:imsizevsBP}.
\end{figure}

In this experiment we test the performance of multiple input sizes while holding the kernel size constant at 5 $\times$ 5. The input sizes varied from $4\times 4$ to $256\times 256$ with a discrete step of $4\times 4$ for the forward propagation experiment and a discrete step of $8\times 8$ for the backward propagation experiment. Different discrete steps are used to highlight the FFT algorithm's best performing size transforms for each propagation.  Each ``input size'' experiment is repeated 10 times and the results are averaged. Figures \ref{fig:imsizevsFP} and \ref{fig:imsizevsBP} show the speed-up over spaceConv vs. input size for the forward and backward propagation, respectively.

For inputs larger than $8 \times 8$ input array sizes (the typical scenario for CNN architectures), OaAconv always outperforms the other methods. The speed-up in the backwards propagation is more significant. The error propagation is enhanced with OaAconv as the kernels are small.

\section{Conclusion}
In this paper we demonstrated that OaAconv can improve the efficiency of CNNs over traditional convolution and convolution via a Hadamard product in the frequency domain. Even though OaAconv must calculate more transforms than FFTconv, the fact that the transforms are smaller outweighs the cost of having to calculate more. In order to have fair comparisons we conducted our experiments on single threads. In practice OaA should be implemented such that each block convolution is computed in parallel, and each FFT is implemented with a GPU FFT library, e.g., cuFFT \cite{nvidia2010cufft} to achieve maximum performance.

In future work we plan to optimize the FFT implementations for small size transforms, and design a CNN entirely in the frequency domain eliminating the bottleneck of the transforms. By creating a CNN that is in the frequency domain, the convolutional layers would only be the Hadamard products. The current challenge to this is to efficiently map an approximation of the non-linear transforms of CNNs (e.g., rectified linear units, sigmoid function, and/or hyperbolic tangent) to the frequency domain.

\section*{Acknowledgements}
We would like to thank Prof. Mateen Rizki of Wright State University for the helpful discussions, and the Air Force Office of Scientific Research (AFOSR) for providing the main funding for this work through LRIR 14Y06COR. Approved for public release, case number: 88ABW-2015-3481

\bibliography{bmvc_final}
\end{document}